\documentclass{article}

\usepackage{arxiv}

\usepackage[utf8]{inputenc} 
\usepackage[T1]{fontenc}    
\usepackage{hyperref}       
\usepackage{url}            
\usepackage{booktabs}       
\usepackage{amsfonts}       
\usepackage{nicefrac}       
\usepackage{microtype}      
\usepackage{lipsum}
\usepackage[english]{babel}
\usepackage[utf8]{inputenc}
\usepackage{algorithm}
\usepackage[noend]{algpseudocode}
\usepackage{multirow}
\usepackage{amsthm}
\usepackage[labelfont=bf,justification=justified,format=plain]{caption}
\usepackage[round]{natbib}

\newtheorem*{remark}{Remark}

\usepackage{amsmath,amsfonts,bm}
\usepackage{mathtools}

\def\eqref#1{equation~\ref{#1}}

\def\1{\bm{1}}

\DeclareMathAlphabet{\mathsfit}{\encodingdefault}{\sfdefault}{m}{sl}
\SetMathAlphabet{\mathsfit}{bold}{\encodingdefault}{\sfdefault}{bx}{n}

\newenvironment{ls}{
\begin{itemize}
  \vspace{-2mm}
  \setlength{\leftskip}{-8mm}
  \setlength{\itemsep}{0pt}
  \setlength{\parskip}{0pt}
  \setlength{\parsep}{0pt}
}{\end{itemize}}

\title{The curious case of developmental BERTology\\
\normalsize{On sparsity, transfer learning, generalization and the brain}}

\author{
  Xin Wang \\
  Cerebras Systems \\
  \texttt{\href{mailto:poincare.disk@gmail.com}{poincare.disk@gmail.com}} \\
}

\begin{document}
\maketitle

\begin{abstract}
    In this essay, we explore a point of intersection between deep learning and neuroscience, through the lens of large language models, transfer learning and network compression. 
    Just like perceptual and cognitive neurophysiology has inspired effective deep neural network architectures which in turn make a useful model for understanding the brain, here we explore how biological neural development might inspire efficient and robust optimization procedures which in turn serve as a useful model for the maturation and aging of the brain. 
\end{abstract}

\keywords{
deep learning 
\and natural language processing
\and BERT
\and network compression
\and sparse neural network
\and neuroscience
\and neural development
}

This essay is written for machine learning researchers and neuroscientists (some jargons in both fields will be used)\footnote{
Originally published on \href{https://towardsdatascience.com/the-curious-case-of-developmental-bertology-d601ec52f69d}{Towards Data Science}.
}. Though it is not intended to be a comprehensive review of literature, we will take a tour through a selection of classic work and new results from a range of topics, in an attempt to develop the following thesis:

\begin{quotation}
    \emph{
    Just like the fruitful interaction between representation learning and perceptual/cognitive neurophysiology, a similar synergy exists between transfer/continual learning, efficient deep learning and developmental neurobiology.
    }
\end{quotation}

Hopefully it would inspire the reader in one way or two, or at the very least, kill some boredom during a global pandemic.

We are going to touch on the following topics through the lens of large language models:
\begin{ls}
    \item How do overparameterized deep neural nets generalize?
    \item How does transfer learning help generalization?
    \item How do we make deep learning computationally efficient in practice?
    \item In tackling these questions, how might deep learning research benefit and benefit from scientific studies of the developing and aging brain?
\end{ls}

\section{A philosophical preamble}

Before we start, it is prudent to say a few words about the brain metaphor, to clarify this author’s position on the issue as it often arises central at debates.

The confluence of deep learning and neuroscience arguably took place as early as the conception of artificial neural nets, because artificial neurons abstract characteristic behaviors of biological ones~\citep{McCulloch1943}. However, the drastically different learning mechanisms and disparities in the kinds of intelligent functions erected a formidable barrier in between the two standing tall for decades. The success of modern deep learning in recent years rekindled another trend of integration, bearing new fruits. In addition to designing AI systems inspired by the brain~\citep[e.g.][]{Hassabis2017}, deep neural nets have recently been proposed to serve as a useful model system to understand how the brain works~\citep[e.g.][]{Richards2019}. The benefits are mutual. Progress is being made in reconciliation of the learning mechanisms~\citep{Lillicrap2020} but, in more than one significant aspect, the intelligence gap obstinately remain~\citep{Marcus2018,Marcus2020}.

Now, for a deep learning researcher or practitioner looking at this mixed landscape today, is a brain analogy \emph{helpful} or \emph{misleading}? It is of course simple to give an answer based on faith, and there are large numbers of believers on both sides. But for now let us not pick a side by belief. Instead, let us evaluate each analogy in its unique context entirely by its practical ramifications: \emph{scientifically}, it is helpful only if it makes experimentally verifiable/falsifiable predictions, and \emph{for engineering}, it is useful only if it generates candidate features that can be subject to solid benchmarking. As such, for all brain analogies we are going to raise in the rest of this essay, however appropriate or farfetched they might seem, we shall look past any prior principles and strive to articulate hypotheses that can guide future scientific and engineering work in practice, either within or beyond the limits of these pages.

\section{The working analogy}

What do we usually think of a deep neural net when likening it to the brain?

For most, the network architecture maps to the gross anatomy of brain areas (such as in a sensory pathway) and their interconnections, i.e. the connectome, units map to neurons or cell assemblies, and connection weights to synaptic strengths. As such, neurophysiology carries out the computation of model inference.

Learning of deep neural nets typically takes place given a pre-defined network architecture, in the form of optimizing an objective function over a training dataset. (A major difficulty lies in the biological plausibility of artificial learning algorithms, a topic we do not touch in this article — here we simply accept the similarity of function despite the differences in mechanism.) Thus, the data-driven learning by optimization is similar to experience-based neural development, i.e. \emph{nurture}, whereas network architecture, and to a large degree initialization and some hyperprameters as well, are genetically programmed as a result of evolution, i.e. \emph{nature}.

\begin{remark}
It should be noted that modern deep net architectures, either implicitly engineered by hand or explicitly optimized through neural architecture search~\citep[NAS, for review see e.g.][]{Wistuba2019}, are also a consequence of data-driven optimization, engendering the inductive bias — the free lunch is paid for by all the unfit that failed to survive natural selection.
\end{remark}

Thanks to the rapid growth of data and computing power, the decade of 2010s saw a Cambrian explosion of deep neural net species, spreading rapidly across the world of machine learning.

\section{BERTology}

The plot thickens as the evolution of modern deep learning produces a cluster of new species in the past two years. They thrive in the continent of natural language understanding (NLU), on fertile deltas of mighty rivers carrying immense computing power, such as the Google and the Microsoft. These remarkable creatures share some key commonalities: they all feature a canonical cortical microcircuitry called the \emph{transformer}~\citep{Vaswani2017}, have rapidly increasing brain volumes setting historic records~\citep[e.g.][]{Shoeybi2019,Microsoft2020,Brown2020} and are often scientifically named after one of the Muppets. But the most prominent common trait of these species crucial to their evolutionary success is the capability of \emph{transfer learning}.

What does this mean? Well, these creatures have a two-stage neural development: a lengthy, self-supervised larval stage called \emph{pre-training} followed by a fast, supervised maturation stage called \emph{fine-tuning}. During self-supervised pre-training, huge corpora of unlabeled text are presented to the subject, who plays with itself by optimizing certain objectives very much similar to solving language quizzes given to human kids, such as completing sentences, filling in missing words, telling logical procession of sentences, and spotting grammatical errors. Then during fine-tuning, a well pre-trained subject can quickly learn to perform a particular language understanding task by supervised training.

Transfer learning’s sweeping conquest of the land of NLU was marked by the advent of bidirectional encoder representations from transformers~\citep[BERT,][]{Devlin2018}. BERT and its variants have advanced the state-of-the-art by a considerable margin. Their remarkable success piqued tremendous interest in the inner workings of these models, creating the study of “BERTology”~\citep[for review, see][]{Rogers2020}. Not unlike neurobiologists, BERTologists stick electrodes into the model brain to record activities for interpretation of the neural code (i.e. activations and attention patterns), make targeted lesions of brain areas (i.e. encoding layers and attention heads) to understand their functions, and study how experiences in early development (i.e. pre-training objectives) contribute to mature behavior (i.e. good performance in NLU tasks).

\section{Network compression}

Meanwhile, in the world of deep learning, multi-stage development (like transfer learning) happens in more animal kingdoms than one. Particularly, in production, one often needs to compress a trained huge neural net into a compact one for efficient deployment.

The practice of network compression derives from one of the very puzzling properties of deep neural nets: \emph{overparameterization helps not only generalization but optimization as well}. That is to say, training a small network is often not only worse than training a large one~\citep[if one can afford to do so of course,][]{Belkin2019}, but also worse than compressing a trained large one to the same small size. In practice, compression can be realized by sparsification (pruning), distillation, etc.

\begin{remark}
It is worth noting that the phenomenon of best sparse network arising from optimizing and then compressing a dense one~\citep[see e.g.][]{Zhu2017,Gale2019} is very much like the developing brain, in which over-produced connections are gradually pruned~\citep{Navlakha2018}.
\end{remark}

The type of multi-stage development in model compression, however, is very different from transfer learning. The two stages of transfer learning see the same model being optimized for different objectives, whereas in model compression, the original model morphs into a different one in order to retain optimality for a same objective. If the former resembles maturation to acquire new skills, then the latter is more like graceful aging without losing already learned skills.

\section{Learning weights vs. learning structures: a duality?}

When a network is compressed, its \emph{structure} often undergoes changes. It could mean either the \emph{network architecture} (e.g. in the case of distillation) or \emph{parameter sparseness} (e.g. in the case of pruning). These structural changes are usually imposed by heuristics or regularizers that constrain the otherwise already effective optimization.

But can \emph{structure} rise above being merely an efficiency constraint and become an effective means for learning? An increasing number of emerging studies seem to suggest so.

One intriguing case is weight-agnostic networks~\citep{Gaier2019}. These jellyfish-like creatures do not have to learn during their lifespan, but still are extremely well adapted to their ecological niches, because evolution did all the heavy lifting in choosing an effective brain structure for them.

Even with a fixed architecture chosen by nature, \emph{learning sparse structure can still be as effective as learning synaptic weights}. Recently, \citet{Ramanujan2019} managed to find sparsified versions of initialized convolutional nets which, if made wide and deep enough, generalize no worse than dense ones undergoing weight training. Theoretical investigations also suggest that sparsification of random weights can be just as effective as optimizing parameters if the model is sufficiently overparameterized~\citep{Malach2020,Ye2020}.

Thus, in the grossly overparameterized regime of modern deep learning, we have in sheath a doubled-edged sword: optimization \emph{of weights} and \emph{of structure}. This is reminiscent of both \emph{synaptic} and \emph{structural plasticity} as mechanisms underlying biological learning and memory~\citep[e.g., see][]{Gage2004,Johansen-Berg2007}.

\begin{figure}[h]
\centering
\includegraphics[scale=0.5]{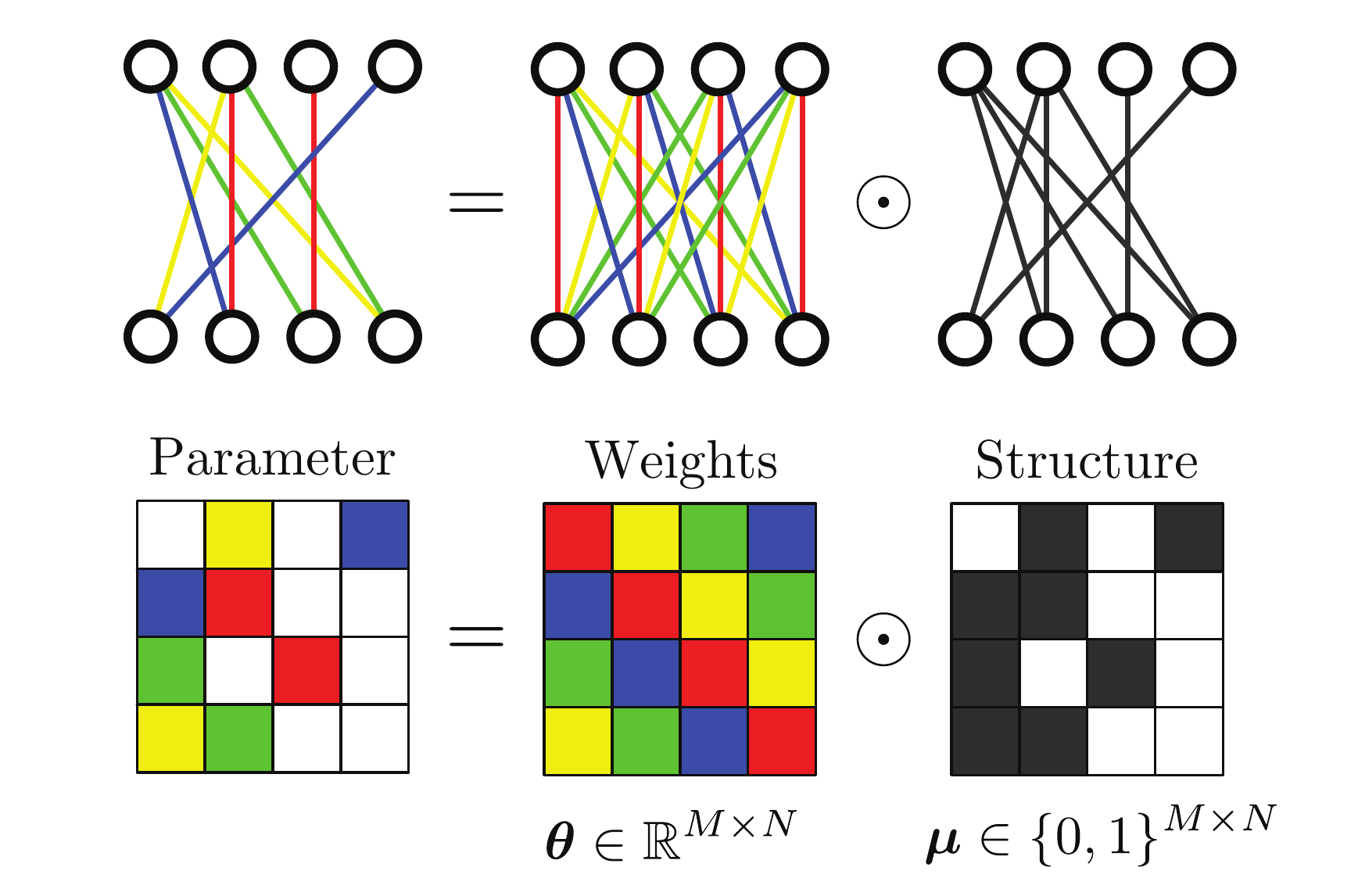}
\caption{The parameter-mask formulation of structural sparseness of model parameters.}
\label{fig:1}
\end{figure}

\begin{figure}[h]
\centering
\includegraphics[scale=0.5]{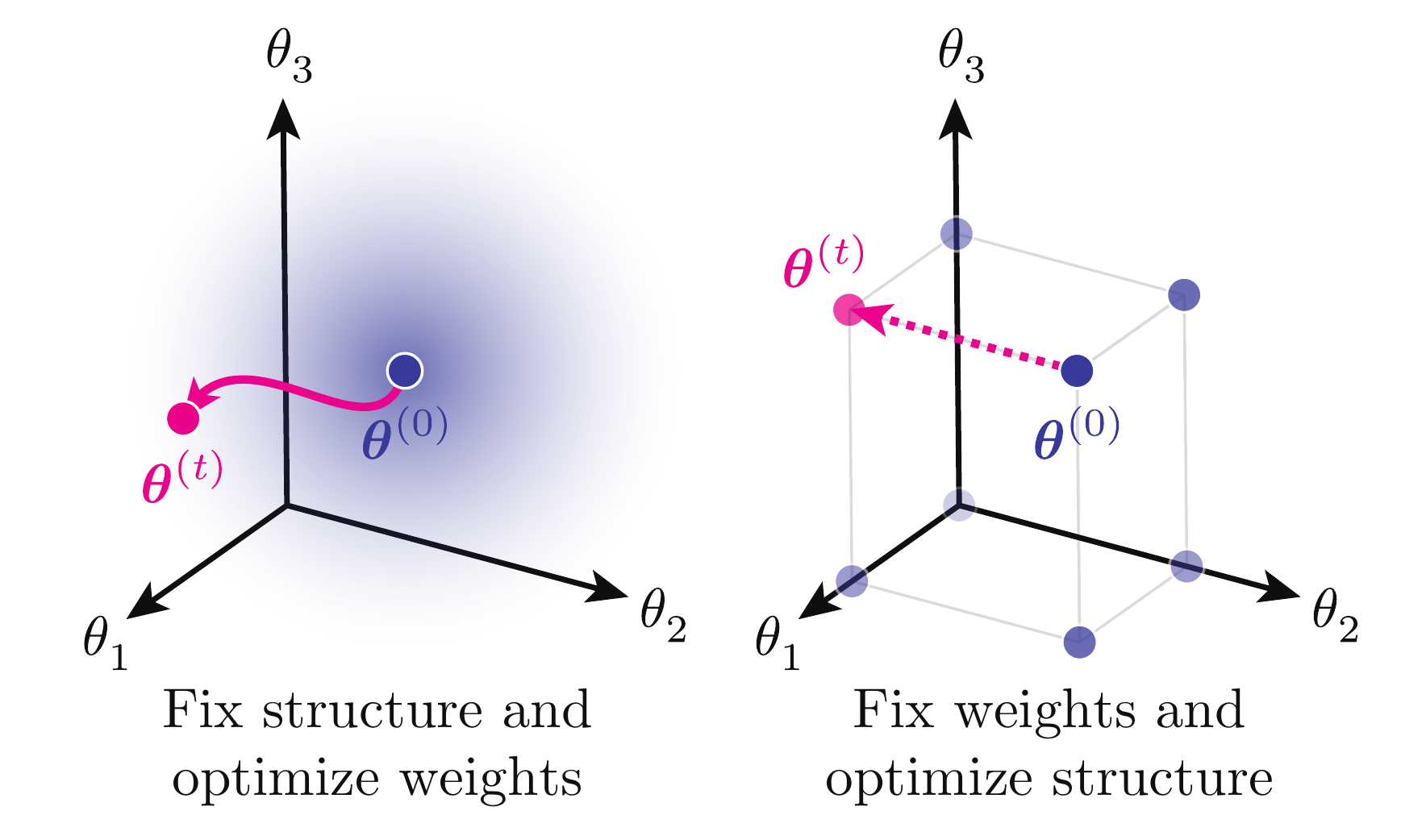}
\caption{Learning weights versus learning structure.}
\label{fig:2}
\end{figure}

\begin{remark}
A formal way of describing parameter sparseness is through the formulation of a parameter mask (Figure~\ref{fig:1}). Learning can be realized either by optimization of continuous weights within a fixed structure, or by optimization of discrete structure given a fixed set of weights (Figure~\ref{fig:2}).
\end{remark}

\section{Fine-tuning by sparsification}

Now that structure, just like weights, can be optimized for learning, can this mechanism be used to make transfer learning better?

Yes, it can indeed. Recently, \citet{Radiya-Dixit2020} made BERT pick up this new gene and evolve to something new. They showed that BERT can be effectively fine-tuned by sparsification of pre-trained weights without changing their values, as demonstrated systematically with the General Language Understanding Evaluation (GLUE) tasks~\citep{Wang2018}.

\begin{figure}[h]
\centering
\includegraphics[scale=0.5]{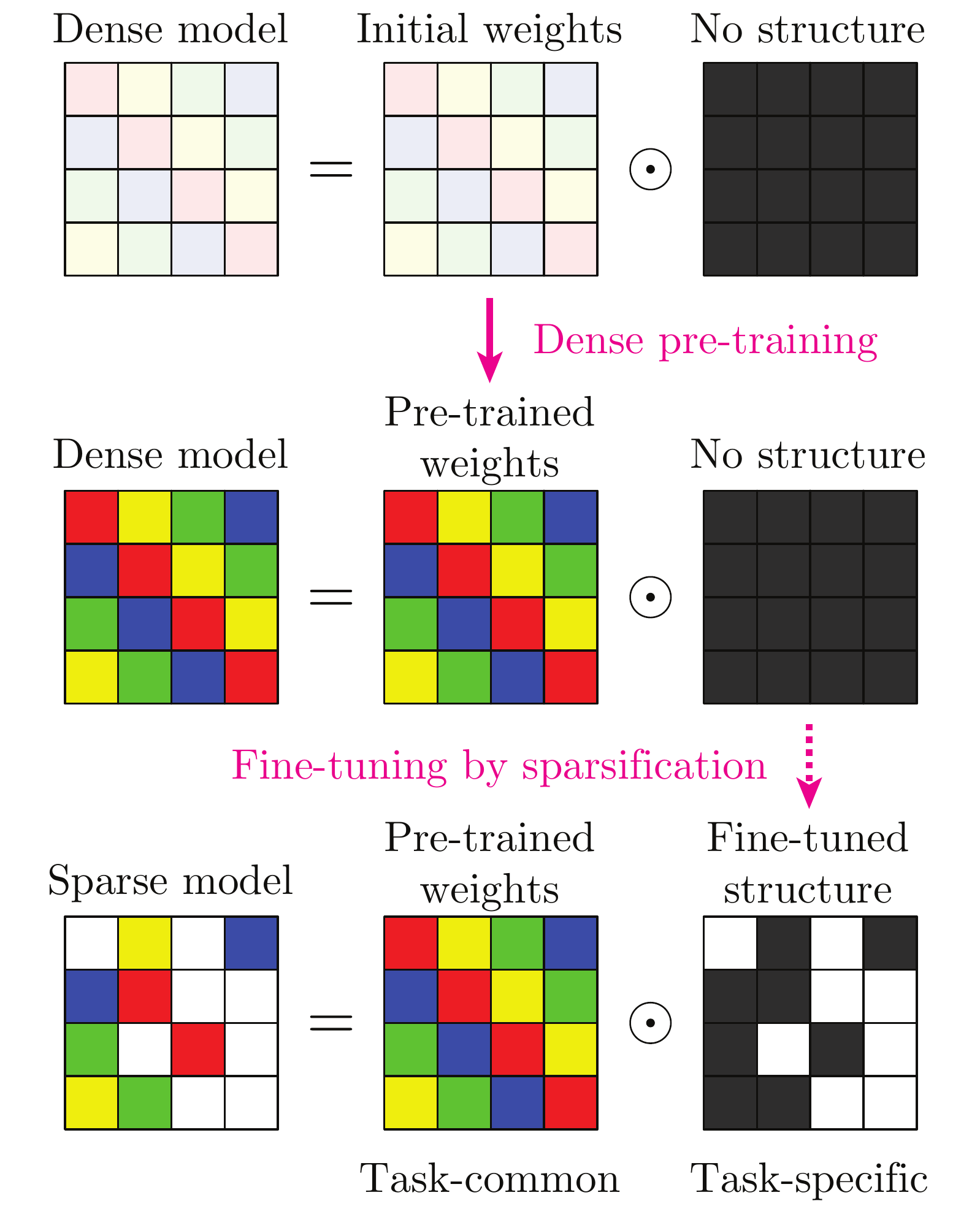}
\caption{Fine-tuning BERT by sparsification~\citep{Radiya-Dixit2020}.}
\label{fig:3}
\end{figure}

\begin{remark}
Note that similar fine-tuning by sparsification has been succesffully applied to computer vision, e.g.~\citet{Mallya2018}. Also take note of existing work sparsifying BERT during pre-training~\citep{Gordon2020}.
\end{remark}

Fine-tuning by sparsification has \emph{favorable practical implications}. On the one hand, pre-trained parameter values remain the same in learning multiple tasks, reducing task-specific parameter storage to only a binary mask; on the other hand, sparsification compresses the model, potentially obviates many “multiply-by-zero-and-accumulate” operations with proper hardware acceleration. One stone kills two birds.

Beyond the practical benefits, however, the possibility of fine-tuning by sparsification brought about a few new opportunities towards a deeper understanding of language pre-training and its potential connections to the biological brain. Let us take a look of them in the next sections.

\section{Winning tickets of a different lottery}

First we study the nature of language pre-training from the perspective of optimization.

It seems that language pre-training meta-learns a good initialization for learning downstream NLU tasks. As \citet{Hao2020} recently showed, pre-trained BERT weights have good task-specific optima that are closer and flatter in loss landscape. This means pre-training makes fine-tuning easier, and the fine-tuned solutions generalize better.

\begin{figure}[h]
\centering
\includegraphics[scale=0.5]{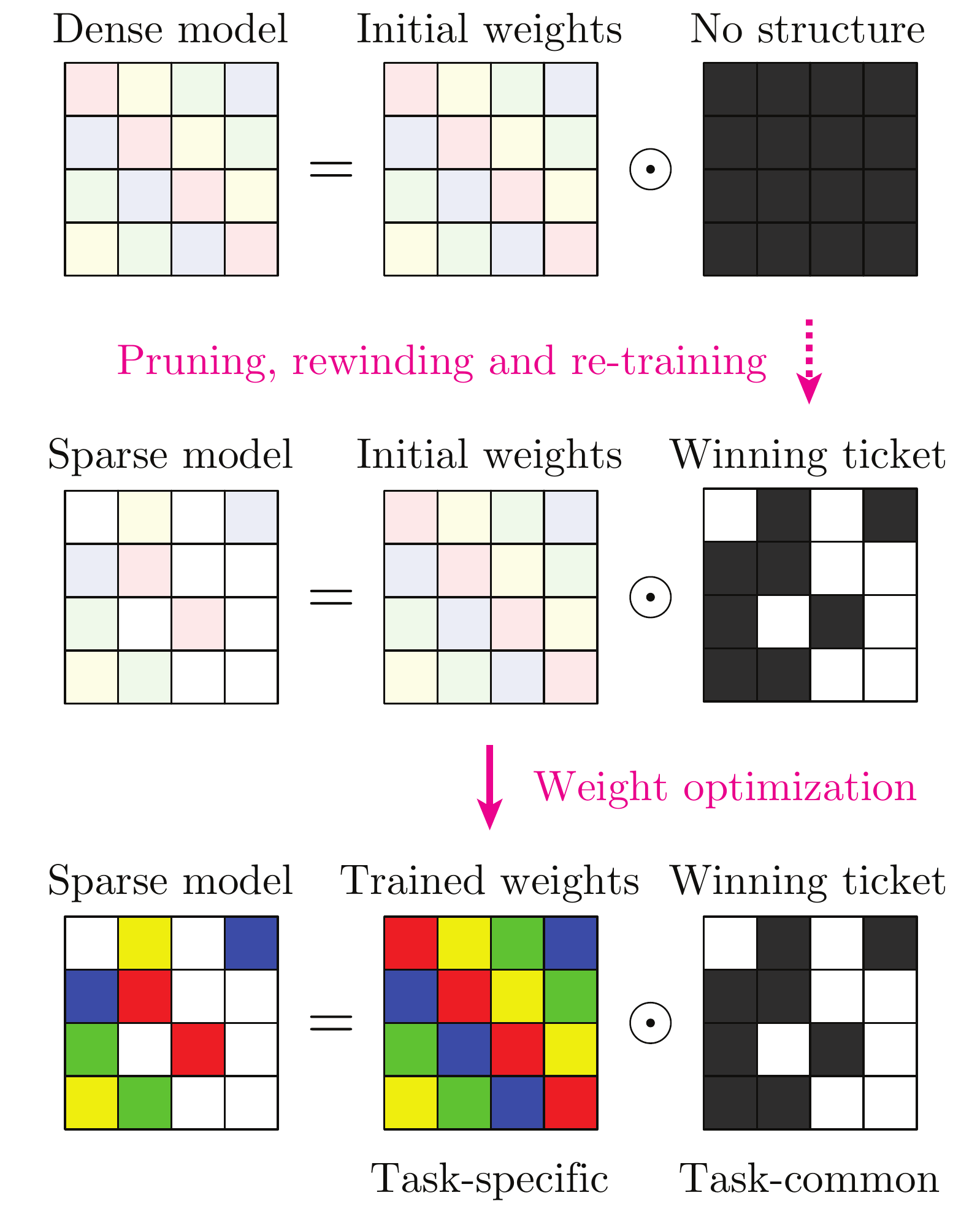}
\caption{The \emph{Frankle-Carbin winning ticket}~\citep{Frankle2018}, cf. fine-tuning by sparsification (Figure~\ref{fig:3}).}
\label{fig:4}
\end{figure}

Similarly, pre-training also makes discovery of fine-tuned sparse subnetworks easier~\citep{Radiya-Dixit2020}. As such, interestingly, pre-trained language models have all the key properties of a “winning lottery ticket” as formulated by \citet{Frankle2018}, but of exactly the complementary kind given the duality of optimizing weights vs. structure (Figures~\ref{fig:3},\ref{fig:4}):

\begin{ls}
    \item The \emph{Frankle-Carbin winning ticket} is a specific sparse structure that facilitates weight optimization. It is sensitive to weight initialization~\citep{Frankle2018}. It is potentially transferable across vision tasks~\citep{Morcos2019}.
    \item A \emph{pre-trained language model} is a specific set of weights that facilitates structural optimization. It is sensitive to structural initialization~\citep{Radiya-Dixit2020}. It is transferable across NLU tasks~\citep{Radiya-Dixit2020}.
\end{ls}

\begin{remark}
Note that the “winning ticket” property of pre-trained BERT is different from the wide-and-deep regime as in \citet{Ramanujan2019}. It remains an open question whether large transformer-based language models, if made sufficiently wide and deep (bound to be astronomically large provided their already huge sizes), might be effectively fine-tuned from random initializations without pre-training.
\end{remark}

Though learning weights of a winning lottery ticket and searching for a subnetwork within pre-trained weights lead to the same outcome — a compact, sparse network that generalizes well, the biological plausibility of the two approaches are drastically different: finding a Frankle-Carbin ticket involves repeated rewinding in time and re-training, a process only possible across multiple biological generations if earlier states could be genetically encoded and then reproduced in the next generation so as to realize rewinding. But weight pre-training followed by structural sparsification are similar to development and aging, all within a single generation. Thus, dense pre-training and sparse fine-tuning might be a useful model for neural development.

\section{Robustness: same function from different structures}

Another uncanny similarity between BERT and the brain is its \emph{structural robustness}.

\begin{figure}[h]
\centering
\includegraphics[scale=0.5]{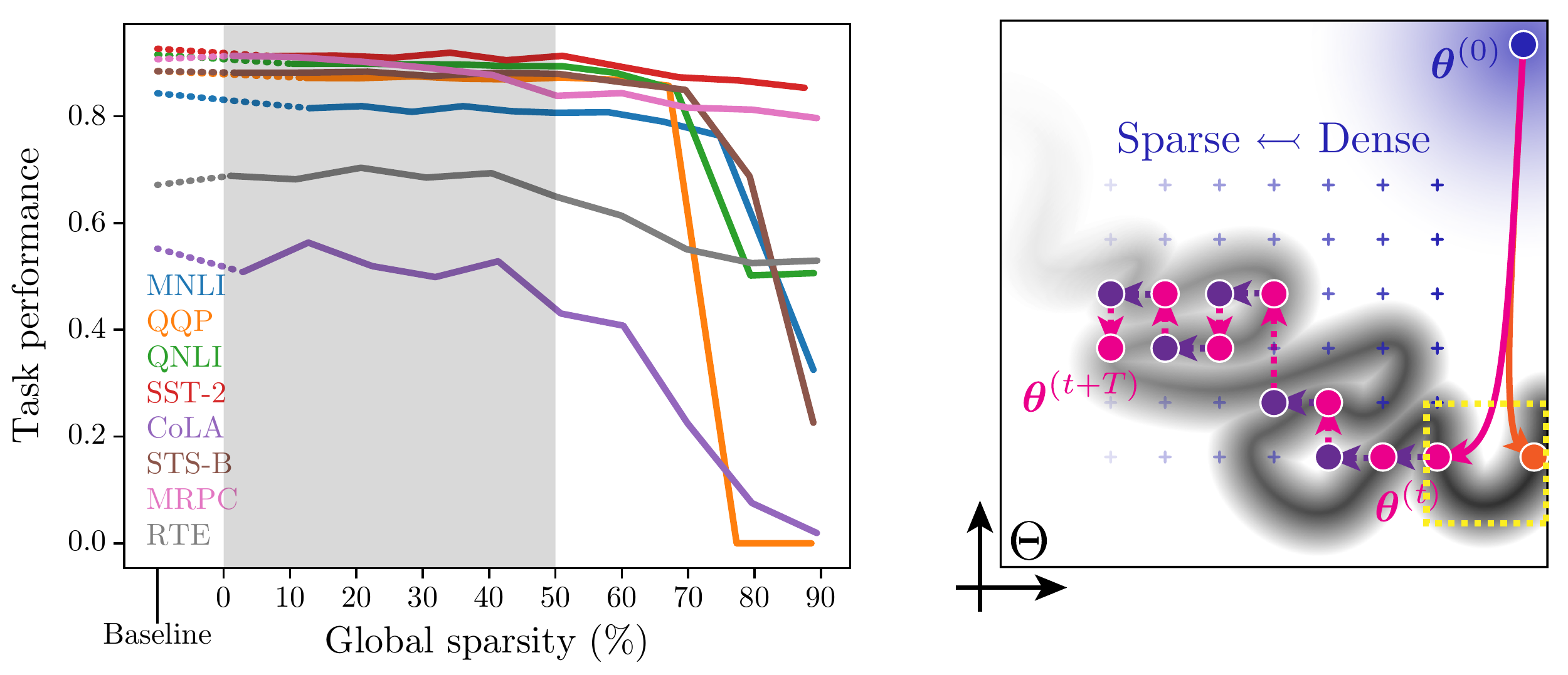}
\caption{
Structural robustness of fine-tuned language models by sparsification. (Left) There exist many good subnetworks of pre-trained BERT that span a wide range of sparsity~\citep[from a few percent to more than half][]{Radiya-Dixit2020}. (Right) A cartoonistic view of the loss landscape during continual sparsification. Dense training (solid magenta and orange arrows) finds low-loss solutions lying on a continuous manifold, dotted yellow box similar to Figure 1 of \citet{Draxler2018}. As long as any structural perturbation by weight elimination (purple dotted arrows and circles) does not deviate far from the low-loss manifold, a quick structural fine-tuning (magenta dotted arrows and circles) can restore optimility, continually. The blue grid represents the discrete set of sparse parameters.
}
\label{fig:5}
\end{figure}

There seems to be an abundance of good subnetworks of pre-trained BERT at a wide range of sparsity levels~\citep{Radiya-Dixit2020}: a typical GLUE task can be learned by eliminating from just a few percent to over half of pre-trained weights, with good sparse solutions exist everywhere in between (Figure~\ref{fig:5}, left). This is reminiscent of structural plasticity at play in the maturing and aging brain — its acquired function remains the same while the underlying structure undergoes continuous changes over time. This is very different from the brittle \emph{point solutions} by traditional engineering.

This phenomenon stems primarily from overparameterization of deep neural nets. In the modern regime of gross overparameterization, optima in the loss landscape are typically high-dimensional continuous non-convex manifolds~\citep{Draxler2018,Fort2019}. This is strangely similar to biology, where identical network behavior can arise from vastly different underlying parameter configurations, forming a non-convex set in the parameter space, e.g. see \citet{Marder2011}.

Here comes the interesting part. Just like the life-long homeostatic adjustment in biology, a similar mechanism might support continual learning in overparameterized deep nets (illustrated in Figure~\ref{fig:5}, right): early-stage learning of dense connections finds a good solution manifold, along which an abundance of good sparse solutions exist; as the network ages, continual and gradual sparsification of the network can be quickly fine-tuned by structural plasticity (like the brain that maintains life-long plasticity).  Having a large, but sparse and plastic brain has functional advantages \citep[e.g.][]{Ahmad2019}.

From the neurobiological perspective, if one accepts \emph{the optimizational hypothesis}~\citep{Richards2019}, then the life-long plasticity must carry out some functional optimization continually during lifespan. Following this logic, neural developmental disorders that arise from this process going awry should essentially be \emph{optimizational diseases}, with etiological characterizations such as bad initialization, unstable optimizer dynamics, etc.

Whether the aforementioned hypothesis holds true for deep neural nets in general, and adequate for them to serve as a good model for neural development and pathophysiology, are open questions for future research.

\section{How much did BERT learn?}

Finally, let us apply some neuroscientific thinking to BERTology.

We ask the question: how much information is stored in \emph{pre-trained} BERT parameters relevant for solving an NLU task? It is not an easy question to answer because sequential changes in parameter values during pre-training and during fine-tuning confound each other.

This limitation is no longer there in the case of BERT fine-tuned by sparsification, where pre-training only learns weight values and fine-tuning only learns structure. To a biologist, it is always good news if two stages of development involve completely different physiological processes, in which case one of them can be used to study the other.

Now let us do exactly this. Let us perturb the pre-trained weight values and study the downstream consequences. For this experiment, we do not make physiological perturbations (such as lesioning attention heads), but a pharmacological one instead: systemic application of a substance that affects every single synapse in the entire brain. This drug is \emph{quantization}. Table~\ref{tab:1} summarizes some preliminary dose-responses: though BERT and related species have developed large brains, it seems knowledge learned during language pre-training might be described by just a few bits per synapse.

\begin{table*}[t]
\caption{
  Fine-tuning by sparsification of quantized pre-trained parameters. Shown are F1 scores of fine-tuned BERT and related models for MRPC, mean $\pm$ S.D. of 3 independent runs. Thanks to \href{https://github.com/huggingface/transformers}{Hugging Face’s \texttt{transformer}}, experiments like this are a breeze.
}
\label{tab:1}
\vspace{-3mm}\centering
\setlength\tabcolsep{4.5pt}
\begin{center}
\begin{small}
\begin{tabular}{l | c c | c c c c}
    \toprule
    \begin{tabular}{@{}l@{}}
        \multirow{2}{*}{
            \begin{tabular}{@{}l@{}}Model\end{tabular}
        }
    \end{tabular} & 
    \begin{tabular}{@{}l@{}}
        \multirow{2}{*}{Fine-tune weights}
    \end{tabular} & 
    \begin{tabular}{@{}l@{}}
        \multirow{2}{*}{Fine-tune structure}
    \end{tabular} & 
    \multicolumn{4}{c}{Fine-tune structure on weights quantized to} \\
    &&&
    \begin{tabular}{@{}l@{}}
        8-bit 
    \end{tabular} & 
    \begin{tabular}{@{}l@{}}
        4-bit  
    \end{tabular} & 
    \begin{tabular}{@{}l@{}}
        2-bit  
    \end{tabular} & 
    \begin{tabular}{@{}l@{}}
        1-bit
    \end{tabular} \\
    \midrule
    \texttt{bert-base} & 
        .8890 $\pm$ .0061 & \bf .9035 $\pm$ .0032 & \bf .9059 $\pm$ .0044 & \bf .8935 $\pm$ .0086 & .8122 $\pm$ .0000 & .5338 $\pm$ .3512 \\
    \texttt{bert-large} & 
        .8968 $\pm$ .0148 & \bf .8996 $\pm$ .0020 & \bf .8995 $\pm$ .0101 & \bf .8977 $\pm$ .0082 & .8122 $\pm$ .0000 & .3792 $\pm$ .1595 \\
    \midrule
    \texttt{xlnet-base} & 
        .8950 $\pm$ .0159 & \bf .9061 $\pm$ .0048 & \bf .9009 $\pm$ .0034 & \bf .8929 $\pm$ .0021 & .8129 $\pm$ .0010 & .8122 $\pm$ .0000 \\
    \texttt{xlnet-large} & 
        .9132 $\pm$ .0051 & \bf .9048 $\pm$ .0038 & \bf .9015 $\pm$ .0312 & \bf .8918 $\pm$ .0152 & .8122 $\pm$ .0000 & .8122 $\pm$ .0000 \\
    \midrule
    \texttt{roberta-base} & 
        .9131 $\pm$ .0057 & \bf .9031 $\pm$ .0025 & \bf .9046 $\pm$ .0095 &     .8459 $\pm$ .0275 & .8122 $\pm$ .0000 & .8122 $\pm$ .0000 \\
    \texttt{roberta-large} & 
        .9158 $\pm$ .0028 & \bf .9186 $\pm$ .0066 & \bf .9124 $\pm$ .0069 &     .8638 $\pm$ .0061 & .8122 $\pm$ .0000 & .7699 $\pm$ .0733 \\
    \midrule
    \texttt{albert-base} & 
        .9008 $\pm$ .0056 & \bf .8984 $\pm$ .0064 & \bf .8962 $\pm$ .0083 & \bf .8945 $\pm$ .0054 & .8086 $\pm$ .0059 & .6906 $\pm$ .0377 \\
    \texttt{albert-large} & 
        .9108 $\pm$ .0021 & \bf .9080 $\pm$ .0029 & \bf .9124 $\pm$ .0094 &     .8251 $\pm$ .0223 & .8121 $\pm$ .0001 & .6682 $\pm$ .0305 \\
    \texttt{albert-xlarge} & 
        .9043 $\pm$ .0025 & \bf .9096 $\pm$ .0062 & \bf .9162 $\pm$ .0064 & \bf .9124 $\pm$ .0060 & .8122 $\pm$ .0000 & .3656 $\pm$ .2633 \\
    \bottomrule
\end{tabular}
\end{small}
\end{center}
\end{table*}

In practice, this means that, since pre-trained weights do not change values during fine-tuning by sparsification, one might only need to store a low-precision integer version of all BERT parameters without any adverse consequences — a significant compression. The upshot: \emph{all you need is a quantized integer version of pre-trained parameters shared across all tasks, with a binary mask fine-tuned for each task}.

\begin{remark}
Note that existing work on quantization of BERT weights quantizes fine-tuned weights~\citep[e.g. Q-BERT,][]{Shen2019} instead of pre-trained weights.
\end{remark}

\section{Epilogue}

Deep neural nets and the brain have obvious differences: at the lowest level, in learning algorithms, and at the highest level, in general intelligence. Nevertheless, profound similarities at intermediate levels have proven beneficial for the advancement of both deep learning and neuroscience.

For instance, perceptual and cognitive \emph{neurophysiology} has already inspired \emph{effective deep network architectures} which in turn make a useful model for understanding the brain. In this essay, we proposed another point of intersection: biological \emph{neural development} might inspire \emph{efficient and robust optimization procedures} which in turn serve as a useful model for maturation and aging of the brain.

\begin{remark}
It should be noted that neural development in the context of traditional connectionism was proposed in the 1990s~\citep[e.g. see][]{RethinkingInnateness}.
\end{remark}

Specifically, we have reviewed some recent results on weight learning and structural learning as complementary means to optimization, and how they, in combination, realize efficient transfer learning in large language models.

As structural learning becomes increasingly important in deep learning, we shall see corresponding hardware accelerators emerge~\citep[e.g. Nvidia’s Ampère architecture supporting sparse weights,][]{Salvator2020}. This is likely to bring about a new wave of architectural diversification of specialized hardware — acceleration of structural learning requires smart data movement adapted to specific computations, a new frontier for exploration.

\bibliography{references}

\end{document}